\documentclass{article} 
\usepackage{iclr2022_conference_arxiv,times}

\usepackage{amsmath,amsfonts,bm,xspace}

\newcommand{\tpc}{{\textsc{TPC}}\xspace}
\newcommand{\cvrl}{{\textsc{CVRL}}\xspace}
\newcommand{\dbc}{{\textsc{DBC}}\xspace}
\newcommand{\pse}{{\textsc{PSE}}\xspace}

\newcommand{\swav}{{\textsc{SwAV}}\xspace}
\newcommand{\rssm}{{\textsc{RSSM}}\xspace}
\newcommand{\planet}{{\textsc{PlaNet}}\xspace}
\newcommand{\dreamer}{{\textsc{Dreamer}}\xspace}
\newcommand{\dreamerb}{{\textsc{DreamerV2}}\xspace}
\newcommand{\dreamerp}{{\textsc{DreamerPro}}\xspace}

\newcommand{\GRU}{{\textsc{GRU}}}

\def\Figref#1{Figure~\ref{#1}}

\def\eqref#1{equation~\ref{#1}}
\def\Eqref#1{Equation~\ref{#1}}
\def\Twoeqref#1#2{Equations \ref{#1} and \ref{#2}}

\def\1{\bm{1}}

\DeclareMathAlphabet{\mathsfit}{\encodingdefault}{\sfdefault}{m}{sl}
\SetMathAlphabet{\mathsfit}{bold}{\encodingdefault}{\sfdefault}{bx}{n}

\def\gJ{{\mathcal{J}}}

\newcommand{\E}{\mathbb{E}}

\newcommand{\softmax}{\mathrm{softmax}}

\newcommand{\KL}{D_{\mathrm{KL}}}

\usepackage{hyperref}
\usepackage{xurl}

\usepackage{graphicx}
\usepackage{booktabs}
\usepackage{adjustbox}

\title{
\dreamerp: Reconstruction-Free \\
Model-Based Reinforcement Learning \\
with Prototypical Representations
}

\author{Fei Deng \\
Rutgers University \\
\texttt{fei.deng@rutgers.edu} \\
\And
Ingook Jang \\
ETRI \\
\texttt{ingook@etri.re.kr} \\
\And
Sungjin Ahn \\
Rutgers University \\
\texttt{sjn.ahn@gmail.com} \\
}

\iclrfinalcopy 
\begin{document}

\maketitle

\begin{abstract}
Top-performing Model-Based Reinforcement Learning (MBRL) agents, such as \dreamer, learn the world model by reconstructing the image observations. Hence, they often fail to discard task-irrelevant details and struggle to handle visual distractions. To address this issue, previous work has proposed to contrastively learn the world model, but the performance tends to be inferior in the absence of distractions. In this paper, we seek to enhance robustness to distractions for MBRL agents. Specifically, we consider incorporating prototypical representations, which have yielded more accurate and robust results than contrastive approaches in computer vision. However, it remains elusive how prototypical representations can benefit temporal dynamics learning in MBRL, since they treat each image independently without capturing temporal structures. To this end, we propose to learn the prototypes from the recurrent states of the world model, thereby distilling temporal structures from past observations and actions into the prototypes. The resulting model, \dreamerp, successfully combines \dreamer with prototypes, making large performance gains on the DeepMind Control suite both in the standard setting and when there are complex background distractions. Code available at \url{https://github.com/fdeng18/dreamer-pro}.
\end{abstract}

\section{Introduction}

Model-Based Reinforcement Learning~\citep[MBRL,][]{rlbook,dyna} provides a solution to many problems in contemporary reinforcement learning. It improves sample efficiency by training a policy through simulations of a learned world model. Learning a world model also provides a way to efficiently represent experience data as general knowledge simulatable and reusable in arbitrary downstream tasks. In addition, it allows accurate and safe decisions via planning. 

Among recent advances in image-based MBRL, \dreamer is particularly notable as the first MBRL model outperforming popular model-free RL algorithms with better sample efficiency in both continuous control~\citep{dreamer} and discrete control~\citep{dreamerv2}. Unlike some previous model-based RL methods~\citep{simple}, it learns a world model that can be rolled out in a compact latent representation space instead of the high-dimensional observation space. Also, policy learning can be done efficiently via backpropagation through the differentiable dynamics model. 

In image-based RL, the key problem is to learn low-dimensional state representation and, in the model-based case, also its forward model. 
Although we can learn such representation directly by maximizing the rewards~\citep{muzero}, it is usually very slow to do this due to the reward sparsity. Instead, it is more practical to introduce auxiliary tasks providing richer learning signal to facilitate representation learning without reward (or with sparse reward)~\citep{horde,unreal}. \dreamer achieves this by learning the representation and the dynamics model in a way to reduce the reconstruction error of the observed sequences.
However, reconstruction-based representation learning has limitations. First, it is computationally expensive to reconstruct the high-dimensional inputs, especially in models like \dreamer that needs to reconstruct long-range videos. Second, it wastes the representation capacity to learn even the visual signals that are irrelevant to the task or unpredictable such as noisy background~\citep{burda2018exploration}. Thus, in MBRL it is of particular interest to realize a version of \dreamer without reconstruction.

Recently, there have been remarkable advances in reconstruction-free representation learning in reinforcement learning~\citep{curl,rad,drq}. The currently dominant approach is via contrastive learning. This approach requires pair-wise comparisons to push apart different instances while pulling close an instance and its augmentation. Therefore, this method usually requires a large batch size (so computationally expensive) to perform accurately and robustly. An alternative is the clustering-based or prototype-based approach~\citep{swav}. By learning a set of clusters represented by prototypes, it replaces the instance-wise comparison by a comparison to the clusters and thereby avoids the problems of contrastive learning. This approach is shown to perform more accurately and robustly in many applications~\citep{swav,dino,protorl} than the contrastive method while also alleviating the need for maintaining a large batch size. The prototype structure can also be used to implement an exploration method~\citep{protorl}.

However, for reconstruction-free MBRL only the contrastive approach like Temporal Predictive Coding~\citep[TPC,][]{tpc} has been proposed so far.
While TPC consistently outperforms \dreamer in the noisy background settings, for standard DeepMind Control suite~\citep{dmc} it showed quite inconsistent results by performing severely worse than \dreamer on some tasks. Therefore, we hypothesize that this inconsistent behavior may be fixed if the robustness and accuracy of the prototypical representations can be realized in MBRL and further improved with the support of temporal information. 

In this paper, we propose a reconstruction-free MBRL agent, called \dreamerp, by combining the prototypical representation learning with temporal dynamics learning. Similar to SwAV~\citep{swav}, by encouraging uniform cluster assignment across the batch, we implicitly pull apart the embeddings of different observations. Additionally, we let the temporal latent state to `reconstruct' the cluster assignment of the observation, thereby relieving the world model from modeling low-level details. We evaluate our model on the standard setting of DeepMind Control suite, and also on a natural background setting, where the background is replaced by natural videos irrelevant to the task. The results show that the proposed model consistently outperforms previous methods.

The contributions of the paper are (1) the first reconstruction-free MBRL agent based on the prototypical representation and its temporal dynamics and (2) the demonstration of the consistently improved accuracy and robustness of the proposed model in comparison to a contrastive reconstruction-free MBRL agent and Dreamer for both standard and natural background DMC tasks.

\section{Preliminaries}

In this section, we briefly introduce the world model and learning algorithms used in $\dreamerb$~\citep{dreamerv2} which our model builds upon. To indicate the general $\dreamer$ framework \citep{dreamer, dreamerv2}, we omit its version number in the rest of the paper.

\subsection{Reconstruction-based world model learning}

$\dreamer$ learns a recurrent state-space model~\citep[$\rssm$,][]{planet} to predict forward dynamics and rewards in partially observable environments. At each time step $t$, the agent receives an image observation $o_t$ and a scalar reward $r_t$ (obtained by previous actions $a_{<t}$). The agent then chooses an action $a_t$ based on its policy. The $\rssm$ models the observations, rewards, and transitions through a probabilistic generative process:
\begin{align}
    p(o_{1:T}, r_{1:T} \mid a_{1:T}) &= \int \prod_{t=1}^{T} p(o_t \mid s_{\leq t}, a_{<t}) \, p(r_t \mid s_{\leq t}, a_{<t}) \, p(s_t \mid s_{<t}, a_{<t}) \  \mathrm{d}s_{1:T}\\
    &= \int \prod_{t=1}^{T} p(o_t \mid h_t, s_t) \, p(r_t \mid h_t, s_t) \, p(s_t \mid h_t) \  \mathrm{d}s_{1:T}\ ,
\end{align}
where the latent variables $s_{1:T}$ are the agent states, and $h_t = \GRU(h_{t-1}, s_{t-1}, a_{t-1})$ is a deterministic encoding of $s_{<t}$ and $a_{<t}$. To infer the agent states from past observations and actions, a variational encoder is introduced:
\begin{align}
    q(s_{1:T} \mid o_{1:T}, a_{1:T}) = \prod_{t=1}^{T} q(s_t \mid s_{<t}, a_{<t}, o_t) = \prod_{t=1}^{T} q(s_t \mid h_t, o_t)\ .
\end{align}
The training objective is to maximize the evidence lower bound (ELBO):
\begin{align} \label{eqn:dreamer_loss}
    \gJ_{\dreamer} = \sum_{t=1}^{T} \E_{q} [\underbrace{\log p(o_t \mid h_t, s_t)}_{\gJ_\mathrm{O}^{t}} + \underbrace{\log p(r_t \mid h_t, s_t)}_{\gJ_\mathrm{R}^{t}} - \underbrace{\KL(q(s_t \mid h_t, o_t) \parallel p(s_t \mid h_t))}_{\gJ_{\mathrm{KL}}^{t}}]\ .
\end{align}

\subsection{Policy learning by latent imagination}

$\dreamer$ interleaves policy learning with world model learning. During policy learning, the world model is fixed, and an actor and a critic are trained cooperatively from the latent trajectories imagined by the world model. Specifically, the imagination starts at each non-terminal state $\hat{z}_t = [h_t, s_t]$ encountered during world model learning. Then, at each imagination step $t' \geq t$, an action is sampled from the actor's stochastic policy: $\hat{a}_{t'} \sim \pi(\hat{a}_{t'} \mid \hat{z}_{t'})$. The corresponding reward $\hat{r}_{t'+1}$ and next state $\hat{z}_{t'+1}$ are predicted by the learned world model. Given the imagined trajectories, the actor improves its policy by maximizing the $\lambda$-return \citep{rlbook, gae} plus an entropy regularizer that encourages exploration, while the critic is trained to approximate the $\lambda$-return through a squared loss.

\section{DreamerPro}

To compute the $\dreamer$ training objective, more specifically $\gJ_\mathrm{O}^{t}$ in \Eqref{eqn:dreamer_loss}, a decoder is required to reconstruct the image observation $o_t$ from the state $z_t = [h_t, s_t]$. Because this reconstruction loss operates in pixel space where all pixels are weighted equally, $\dreamer$ tends to allocate most of its capacity to modeling complex visual patterns that cover a large pixel area (e.g., backgrounds). This leads to poor task performance when those visual patterns are task irrelevant, as shown in previous work \citep{tpc}.

Fortunately, during policy learning, what we need is accurate reward and next state prediction, which are respectively encouraged by $\gJ_\mathrm{R}^{t}$ and $\gJ_{\mathrm{KL}}^{t}$. In other words, the decoder is not required for policy learning. The main purpose of having the decoder and the associated loss $\gJ_\mathrm{O}^{t}$, as shown in $\dreamer$, is to learn meaningful representations that cannot be obtained by $\gJ_\mathrm{R}^{t}$ and $\gJ_{\mathrm{KL}}^{t}$ alone.

The above observations motivate us to improve robustness to visual distractions by replacing the reconstruction-based representation learning in $\dreamer$ with reconstruction-free methods. For this, we take inspiration from recent developments in self-supervised image representation learning, which can be divided into contrastive \citep{cpc, simclr, moco} and non-contrastive \citep{byol, swav} methods. We prefer non-contrastive methods as they can be applied to small batch sizes. This can speed up both world model learning and policy learning (in wall clock time). Therefore, we propose to combine $\dreamer$ with the prototypical representations used in $\swav$ \citep{swav}, a top-performing non-contrastive representation learning method. We name the resulting model $\dreamerp$, and provide the model description in the following.

\begin{figure}[t]
\centering
\includegraphics[width=0.9\textwidth]{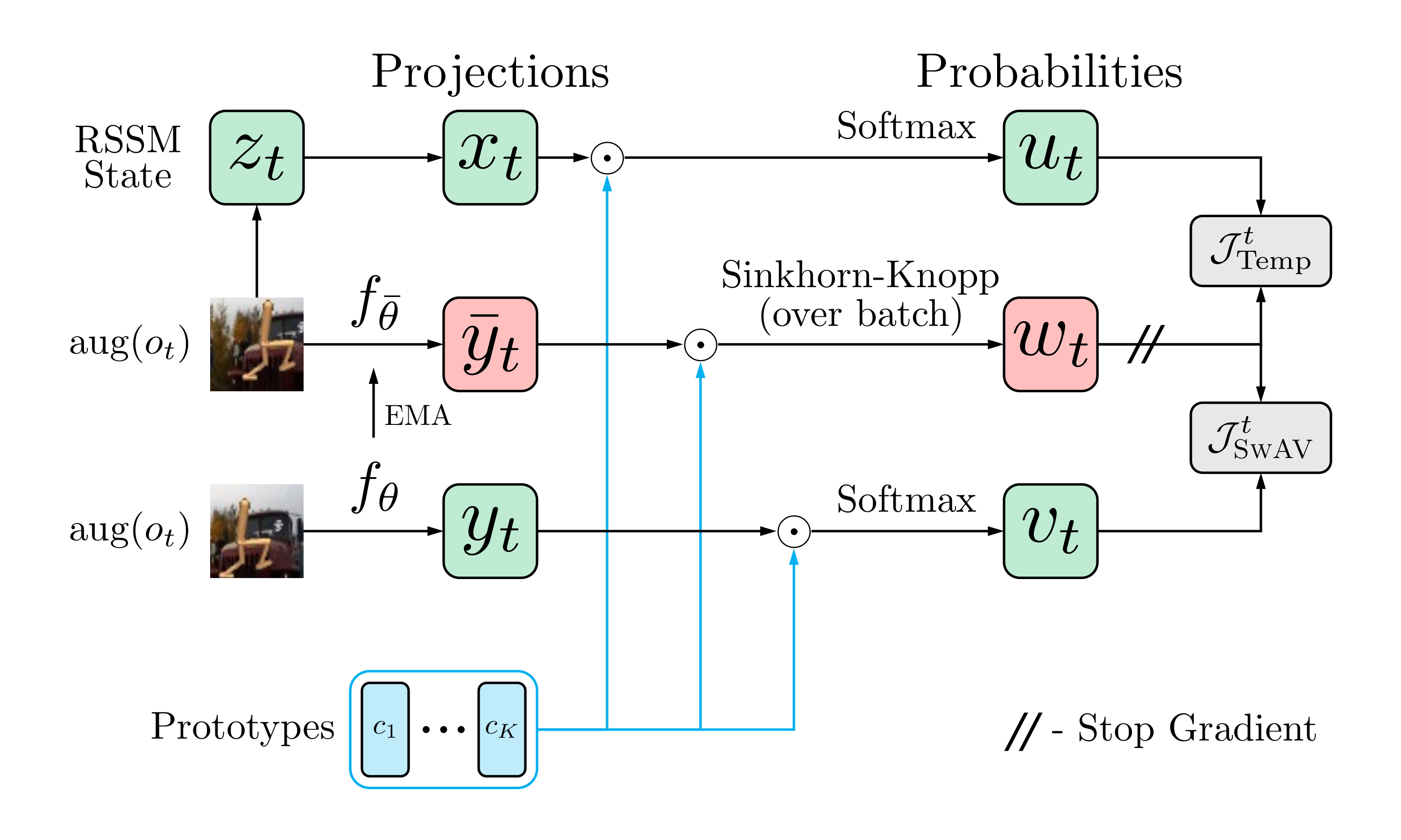}
\caption{\dreamerp learns the world model through online clustering, eliminating the need for reconstruction. At each time step $t$, it first compares the observation to a set of trainable prototypes $\{c_1, \dots, c_K\}$ to obtain the target cluster assignment $w_t$. Then, it predicts this target from both the world model state $z_t$ and another augmented view of the observation. The predictions are improved by optimizing the two objective terms, $\gJ_\mathrm{Temp}^{t}$ and $\gJ_\swav^{t}$, respectively, where the first term crucially distills temporal structures from $z_t$ into the prototypes.}
\label{fig:model}
\end{figure}

$\dreamerp$ uses the same policy learning algorithm as $\dreamer$, but learns the world model without reconstructing the observations. This is achieved by clustering the observation into a set of $K$ trainable prototypes $\{c_1, \dots, c_K\}$, and then predicting the cluster assignment from the state as well as an augmented view of the observation. See \Figref{fig:model} for an illustration.

Concretely, given a sequence of observations $o_{1:T}$ sampled from the replay buffer, we obtain two augmented views $o_{1:T}^{(1)}, o_{1:T}^{(2)}$ by applying random shifts \citep{rad, drq} with bilinear interpolation \citep{drqv2}. We ensure that the augmentation is consistent across time steps. Each view $i \in \{1, 2\}$ is fed to the $\rssm$ to obtain the states $z_{1:T}^{(i)}$. To predict the cluster assignment from $z_t^{(i)}$, we first apply a linear projection followed by $\ell_2$-normalization to obtain a vector $x_t^{(i)}$ of the same dimension as the prototypes, and then take a softmax over the dot products of $x_t^{(i)}$ and all the prototypes:
\begin{align} \label{eqn:pred_state}
    (u_{t,1}^{(i)}, \dots, u_{t,K}^{(i)}) = \softmax\left(\frac{x_t^{(i)} \cdot c_1}{\tau}, \dots, \frac{x_t^{(i)} \cdot c_K}{\tau}\right)\ .
\end{align}
Here, $u_{t,k}^{(i)}$ is the predicted probability that state $z_t^{(i)}$ maps to cluster $k$, $\tau$ is a temperature parameter, and the prototypes $\{c_1, \dots, c_K\}$ are also $\ell_2$-normalized.

Analogously, to predict the cluster assignment from an augmented observation $o_t^{(i)}$, we feed it to a convolutional encoder (shared with the $\rssm$), apply a linear projection followed by $\ell_2$-normalization, and obtain a vector $y_t^{(i)}$. We summarize this process as: $y_t^{(i)} = f_\theta (o_t^{(i)})$, where $\theta$ collectively denotes the parameters of the convolutional encoder and the linear projection layer. The prediction probabilities are again given by a softmax:
\begin{align} \label{eqn:pred_obs}
    (v_{t,1}^{(i)}, \dots, v_{t,K}^{(i)}) = \softmax\left(\frac{y_t^{(i)} \cdot c_1}{\tau}, \dots, \frac{y_t^{(i)} \cdot c_K}{\tau}\right)\ ,
\end{align}
where $v_{t,k}^{(i)}$ is the predicted probability that observation $o_t^{(i)}$ maps to cluster $k$.

To obtain the targets for the above two predictions (i.e., \Twoeqref{eqn:pred_state}{eqn:pred_obs}), we apply the Sinkhorn-Knopp algorithm \citep{sinkhorn} to the cluster assignment scores computed from the output of a momentum encoder $f_{\bar{\theta}}$ \citep{moco, byol, dino}, whose parameters $\bar{\theta}$ are updated using the exponential moving average of $\theta$: $\bar{\theta} \leftarrow (1 - \eta)\bar{\theta} + \eta \theta$. For each observation $o_t^{(i)}$, the scores are given by the dot products $(\bar{y}_t^{(i)} \cdot c_1, \dots, \bar{y}_t^{(i)} \cdot c_K)$, where $\bar{y}_t^{(i)} = f_{\bar{\theta}} (o_t^{(i)})$ is the momentum encoder output. The Sinkhorn-Knopp algorithm is applied to the two augmented batches $\{o_{1:T}^{(1)}\}, \{o_{1:T}^{(2)}\}$ separately to encourage uniform cluster assignment within each augmented batch and avoid trivial solutions. We specifically choose the number of prototypes $K = B \times T$, where $B$ is the batch size, so that the observation embeddings are implicitly pushed apart from each other. The outcome of the Sinkhorn-Knopp algorithm is a set of cluster assignment targets $(w_{t,1}^{(i)}, \dots, w_{t,K}^{(i)})$ for each observation $o_t^{(i)}$.

Now that we have the cluster assignment predictions and targets, the representation learning objective is simply to maximize the prediction accuracies:
\begin{align}
    \gJ_\swav^{t} &= \frac{1}{2} \sum_{k=1}^{K} \left(w_{t,k}^{(1)} \log v_{t,k}^{(2)} + w_{t,k}^{(2)} \log v_{t,k}^{(1)}\right)\ ,\\
    \gJ_\mathrm{Temp}^{t} &= \frac{1}{2} \sum_{k=1}^{K} \left(w_{t,k}^{(1)} \log u_{t,k}^{(1)} + w_{t,k}^{(2)} \log u_{t,k}^{(2)}\right)\ .
\end{align}
Here, $\gJ_\swav^{t}$ improves prediction from an augmented view. This is the same loss as used in $\swav$ \citep{swav}, and is shown to induce useful features for static images. However, it ignores the temporal structure which is crucial in reinforcement learning. Hence, we add a second term, $\gJ_\mathrm{Temp}^{t}$, that improves prediction from the state of the same view. This has the effect of making the prototypes close to the states that summarize the past observations and actions, thereby distilling temporal structure into the prototypes. From another perspective, $\gJ_\mathrm{Temp}^{t}$ is similar to $\gJ_\mathrm{O}^{t}$ in the sense that we are now `reconstructing' the cluster assignment of the observation instead of the observation itself. This frees the world model from modeling complex visual details, allowing more capacity to be devoted to task-relevant features.

The overall world model learning objective for $\dreamerp$ can be obtained by replacing $\gJ_\mathrm{O}^{t}$ in \Eqref{eqn:dreamer_loss} with $\gJ_\swav^{t} + \gJ_\mathrm{Temp}^{t}$:
\begin{align} \label{eqn:dreamerp_loss}
    \gJ_{\dreamerp} = \sum_{t=1}^{T} \E_{q} [\gJ_\swav^{t} + \gJ_\mathrm{Temp}^{t} + \gJ_\mathrm{R}^{t} - \gJ_{\mathrm{KL}}^{t}]\ ,
\end{align}
where $\gJ_\mathrm{R}^{t}$ and $\gJ_{\mathrm{KL}}^{t}$ are now averaged over the two augmented views.

\section{Experiments}

\textbf{Environments.}
We evaluate our model and the baselines on six image-based continuous control tasks from the DeepMind Control (DMC) suite \citep{dmc}. We choose the set of tasks based on those considered in $\planet$ \citep{planet}. Specifically, we replace Cartpole Swingup and Walker Walk with their more challenging counterparts, Cartpole Swingup Sparse and Walker Run, and keep the remaining tasks. In addition to the standard setting, we also consider a natural background setting \citep{dbc, tpc}, where the background is replaced by task-irrelevant natural videos randomly sampled from the `driving car' class in the Kinetics 400 dataset \citep{kinetics}. Following $\tpc$ \citep{tpc}, we use two separate sets of background videos for training and evaluation. Hence, the natural background setting tests generalization to unseen distractions. We note that the recently released Distracting Control Suite~\citep[DCS,][]{dcs} serves a similar purpose. However, the background distractions in DCS seem less challenging, as there are fewer videos and the ground plane is made visible for most tasks. In our preliminary experiments, our model and all the baselines achieved close to zero returns on Cartpole Swingup Sparse in the natural background setting. We therefore switch back to Cartpole Swingup in this setting.

\textbf{Baselines.}
Our main baselines are $\dreamer$ \citep{dreamerv2} and $\tpc$ \citep{tpc}, the state-of-the-art for reconstruction-based and reconstruction-free model-based reinforcement learning, respectively. In particular, $\tpc$ has shown better performance than $\cvrl$ \citep{cvrl} and $\dbc$ \citep{dbc} on the same datasets. The recently proposed $\pse$ \citep{pse} has demonstrated impressive results on DCS. However, it is only shown to work in the model-free setting and requires a pretrained policy, while our model learns both the world model and the policy from scratch.

\textbf{Implementation details.}
We implement our model based on a newer version of $\dreamer$\footnote{\url{https://github.com/danijar/dreamerv2/tree/e783832f01b2c845c195587158c4e129edabaebb}}, while the official implementation of $\tpc$\footnote{\url{https://github.com/VinAIResearch/TPC-tensorflow}} is based on an older version. For fair comparison, we re-implement $\tpc$ based on the newer version. We adopt the default values for the $\dreamer$ hyperparameters, except that we use continuous latents and \texttt{tanh\_normal} as the distribution output by the actor. We find these changes improve $\dreamer$'s performance in the standard DMC, and therefore use these values for all models in both the standard and the natural background setting. Following $\tpc$, we increase the weight of the reward loss $\gJ_\mathrm{R}^{t}$ to $1000$ for all models in the natural background setting to further encourage extraction of task-relevant information. While in the original $\tpc$, this weight is chosen separately for each task from $\{100, 1000\}$, we find the weight of $1000$ works consistently better in our re-implementation, which also obtains better results than reported in the original paper.

\begin{figure}[t]
\centering
\includegraphics[width=0.9\textwidth]{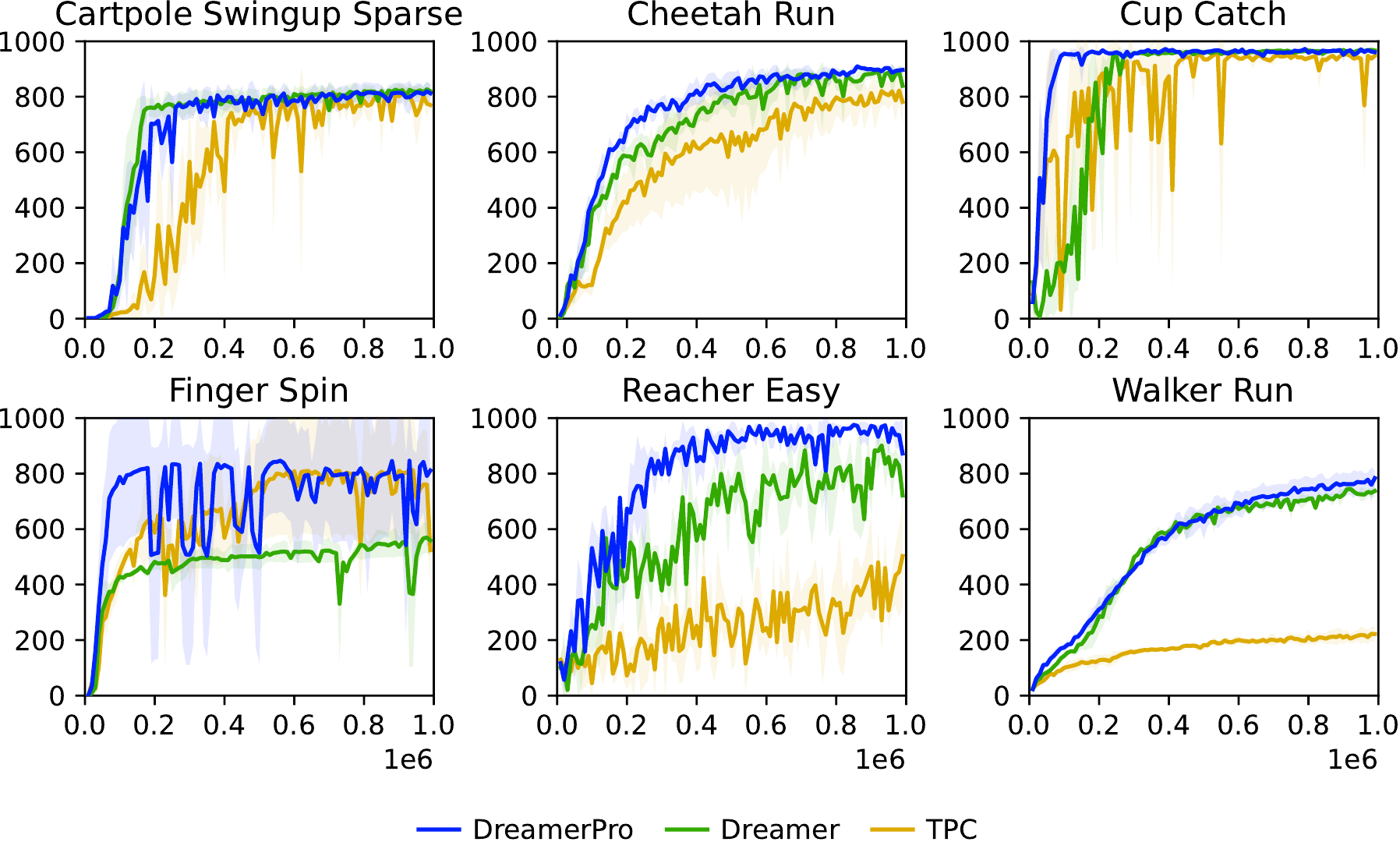}
\caption{Performance curves in standard DMC. While $\tpc$ underperforms $\dreamer$ on most tasks, $\dreamerp$ greatly outperforms $\dreamer$ on Finger Spin and Reacher Easy, achieves better data efficiency on Cup Catch, and is comparable or better than $\dreamer$ on other tasks.}
\label{fig:dmc}
\end{figure}

\begin{table}[t]
	\centering
	\caption{Final performance in standard DMC.}
	\label{tbl:dmc}
	\vskip 0.1in
	\begin{adjustbox}{max width=\textwidth}
  	\begin{tabular}{cccc}
      \toprule
      Task & $\dreamer$ & $\tpc$ & $\dreamerp$ \\
      \midrule
  	  \multicolumn{1}{l}{Cartpole Swingup Sparse} & $\mathbf{820\pm23}$ & $770\pm9$ & $\mathbf{813\pm32}$ \\
  	  \multicolumn{1}{l}{Cheetah Run} & $840\pm74$ & $782\pm82$ & $\mathbf{897\pm8}$ \\
      \multicolumn{1}{l}{Cup Catch} & $\mathbf{967\pm3}$ & $948\pm7$ & $\mathbf{961\pm10}$ \\
      \multicolumn{1}{l}{Finger Spin} & $559\pm54$ & $524\pm127$ & $\mathbf{811\pm232}$ \\
      \multicolumn{1}{l}{Reacher Easy} & $721\pm51$ & $503\pm185$ & $\mathbf{873\pm127}$ \\
      \multicolumn{1}{l}{Walker Run} & $737\pm26$ & $222\pm29$ & $\mathbf{784\pm28}$ \\
      \midrule
      \multicolumn{1}{l}{Mean and STD of NDB ($\downarrow$)} & $0.101\pm0.12$ & $0.284\pm0.272$ & $\mathbf{0.002\pm0.003}$ \\
      \bottomrule
  	\end{tabular}
  	\end{adjustbox}
\end{table}

\textbf{Evaluation protocol.}
For each task, we train each model for $1$M environment steps (equivalent to $500$K actor steps, as the action repeat is set to $2$). The evaluation return is computed every $10$K steps, and averaged over $10$ episodes. In all figures and tables, the mean and standard deviation are computed from $3$ independent runs. In addition to the episode returns, we also report the Normalized Distance to the Best (NDB). Given the performance $x_{\tau,\alpha}$ of an algorithm $\alpha$ on a task $\tau$, the NDB $\beta_{\tau,\alpha}$ is defined as:
\begin{align}
    \beta_{\tau,\alpha} = \frac{\max_{\hat{\alpha}}(x_{\tau,\hat{\alpha}}) - x_{\tau,\alpha}}{\max_{\hat{\alpha}}(x_{\tau,\hat{\alpha}})} \ .
\end{align} 
We report the mean and variance of $\beta_{\tau,\alpha}$ for each algorithm $\alpha$ over all tasks $\tau$, which indicate the bestness and consistency of algorithm $\alpha$, respectively. An optimal algorithm that consistently achieves the best performance across all tasks will have zero mean and zero variance.

\subsection{Performance in standard DMC}

We show the performance curves in \Figref{fig:dmc} and the final performance in Table~\ref{tbl:dmc} for the standard setting. In contrast to $\tpc$ which underperforms $\dreamer$ on most tasks (most severely on Reacher Easy and Walker Run), $\dreamerp$ achieves comparable or even better performance than $\dreamer$ on all tasks. Notably, $\dreamerp$ outperforms $\dreamer$ by a large margin on Finger Spin and Reacher Easy, and demonstrates better data efficiency on Cup Catch. We notice a large variance in $\dreamerp$'s performance on Finger Spin. Further investigation reveals that $\dreamerp$ learned close to optimal behavior (with average episode returns above $950$) on two of the seeds, while converged to a suboptimal behavior (with average episode returns around $500$) on the other seed. The low variance of \dreamer indicates that it hardly achieved close to optimal behavior. Our results suggest for the first time that prototypical representations (and reconstruction-free representation learning in general) can be beneficial to MBRL even in the absence of strong visual distractions.

\subsection{Performance in natural background DMC}

\begin{figure}[t]
\centering
\includegraphics[width=0.9\textwidth]{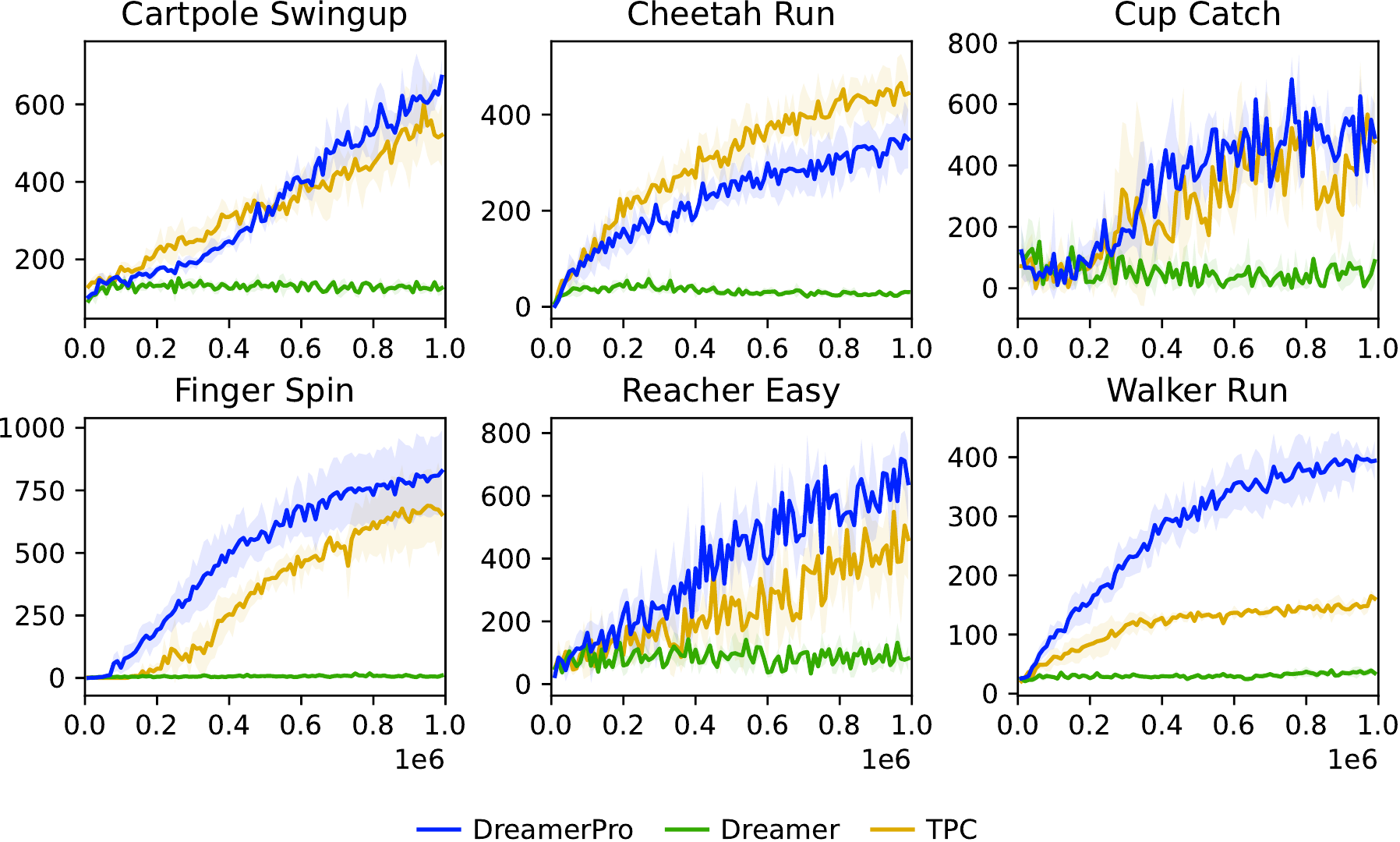}
\caption{Performance curves in natural background DMC. $\dreamerp$ significantly outperforms $\tpc$ on Cartpole Swingup, Finger Spin, Reacher Easy, and Walker Run, while $\dreamer$ completely fails on all tasks.}
\label{fig:nat}
\end{figure}

\begin{table}[t]
	\centering
	\caption{Final performance in natural background DMC.}
	\label{tbl:nat}
	\vskip 0.1in
	\begin{adjustbox}{max width=\textwidth}
  	\begin{tabular}{cccc}
      \toprule
      Task & $\dreamer$ & $\tpc$ & $\dreamerp$ \\
      \midrule
  	  \multicolumn{1}{l}{Cartpole Swingup} & $126\pm16$ & $521\pm80$ & $\mathbf{671\pm42}$ \\
  	  \multicolumn{1}{l}{Cheetah Run} & $30\pm2$ & $\mathbf{444\pm35}$ & $349\pm61$ \\
      \multicolumn{1}{l}{Cup Catch} & $88\pm73$ & $477\pm175$ & $\mathbf{493\pm109}$ \\
      \multicolumn{1}{l}{Finger Spin} & $10\pm1$ & $655\pm133$ & $\mathbf{826\pm162}$ \\
      \multicolumn{1}{l}{Reacher Easy} & $82\pm39$ & $462\pm130$ & $\mathbf{641\pm123}$ \\
      \multicolumn{1}{l}{Walker Run} & $35\pm4$ & $161\pm6$ & $\mathbf{394\pm33}$ \\
      \midrule
      \multicolumn{1}{l}{Mean and STD of NDB ($\downarrow$)} & $0.890\pm0.063$ & $0.222\pm0.237$ & $\mathbf{0.036\pm0.096}$ \\
      \bottomrule
  	\end{tabular}
  	\end{adjustbox}
\end{table}

\Figref{fig:nat} and Table~\ref{tbl:nat} respectively show the performance curves and final evaluation returns obtained by all models in the natural background setting. $\dreamer$ completely fails on all tasks, showing the inability of reconstruction-based representation learning to deal with complex visual distractions. In contrast, $\dreamerp$ achieves the best performance on 5 out of 6 tasks, with large performance gains from $\tpc$ on Cartpole Swingup, Finger Spin, Reacher Easy, and Walker Run. These results indicate that the advantage of prototypical representations over contrastive learning in computer vision can indeed be transferred to MBRL for better robustness to visual distractions.

\subsection{Ablation study}

\begin{figure}[t]
\centering
\includegraphics[width=\textwidth]{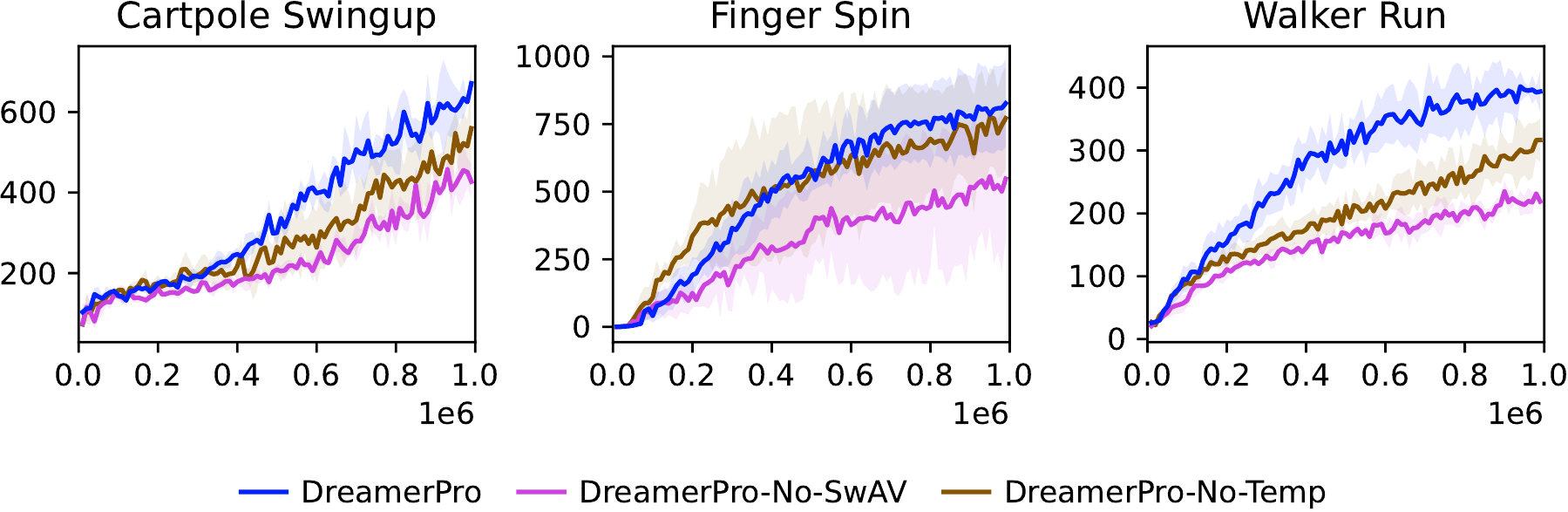}
\caption{Ablation study. Both $\gJ_\swav^{t}$ and $\gJ_\mathrm{Temp}^{t}$ are necessary for achieving good performance.}
\label{fig:ablation}
\end{figure}

We now show the individual effect of the two loss terms, $\gJ_\swav^{t}$ and $\gJ_\mathrm{Temp}^{t}$, in \Figref{fig:ablation}. Here, each of the ablated versions, DreamerPro-No-SwAV and DreamerPro-No-Temp, removes one of the loss terms. We did not investigate removing both terms, as its failure has been shown in $\dreamer$ \citep{dreamer}. We observe that both terms are necessary for achieving good performance. In particular, naively combining $\swav$ with $\dreamer$ (i.e., DreamerPro-No-Temp) leads to inferior performance, as it ignores the temporal structure. On the other hand, $\gJ_\mathrm{Temp}^{t}$ alone is not sufficient to provide meaningful cluster assignment targets and learning signals for the convolutional encoder.

\section{Related work}

\textbf{Self-supervised representation learning for static images.}
Recent works in self-supervised learning have shown its effectiveness in learning representations from high-dimensional data. CPC~\citep{cpc} learns representations by maximizing the mutual information between the encoded representations and its future prediction using noise-contrastive estimation. SimCLR~\citep{simclr} shows that the contrastive data can be generated using the data in the training mini-batch by applying random augmentations. MoCo~\citep{moco}, on the other hand, improves the contrastive training by generating the representations from a momentum encoder instead of the trained network. Despite the success in some tasks, one weakness of the contrastive approaches is that it require the model to compare a larger amount of samples, which demands large batch sizes or memory banks. To address this problem, some works propose to learn the image representations without discriminating between samples. Particularly, BYOL~\citep{byol} introduces a momentum encoder to provide target representations for the training network. SwAV~\citep{swav} proposes to learn the embeddings by matching them to a set of learned clusters. DINO~\citep{dino} replaces the clusters in SwAV with categorical heads and uses the centering and sharpening technique to prevent representations collapsing. Unlike our model, these works treat each image independently and ignore the temporal structure of the environment, which is crucial in learning the forward dynamics and policy in MBRL.

\textbf{Representation learning for model-free reinforcement learning.}
It has been shown that adopting data augmentation techniques like random shifts in the observation space enables robust learning from pixel input in any model-free reinforcement learning algorithm \citep{rad, drq, drqv2}. Recent works have also shown that self-supervised representation learning techniques can bring significant improvement to reinforcement learning methods. For example, CURL~\citep{curl} performs contrastive learning along with off-policy RL algorithms and shows that it signiﬁcantly improves sample-efﬁciency and model performance over pixel-based methods. Other works aim to improve the representation learning quality by combining temporal prediction models in the representation learning process \citep{spr, sgi, atc, protorl, pbl, simcore}. However, the main purpose of the temporal prediction models in these works is mainly to obtain the abstract representations of the observations, and they are not shown to support long-horizon imagination. 

\textbf{Model-based reinforcement learning with reconstruction.} Model based reinforcement learning from raw pixel data can learn the representation space by minimizing the observation reconstruction loss. World Models \citep{worldmodels} learn the latent dynamics of the environment in a two-stage process to evolve their linear controllers in imagination. SOLAR \citep{solar} models the dynamics as time-varying linear-Gaussian and solves robotic tasks via guided policy search. Dreamer \citep{dreamer} jointly learns the RSSM and latent state space from observation reconstruction loss. DeepMDP \citep{deepmdp} also propose a latent dynamics model-based method that uses bisimulation metrics and reconstruction loss in Atari. However, reconstruction based methods are susceptible to noise and objects irrelevant to the task in the environment \citep{tpc}. Furthermore, in a few cases, the latent representation fails to reconstruct small task-relevant objects in the environment \citep{dreaming}. 

\textbf{Reinforcement learning under visual distractions.}
A large body of works on robust representation learning focuses on contrastive objectives. For example, CVRL \citep{cvrl} proposes to learn representations from complex observations by maximizing the mutual information between an image and its corresponding embedding using contrastive objectives. However, the learning objective of CVRL encourages the representation model to learn as much information as possible, including task-irrelevant information. Dreaming \citep{dreaming} and TPC \citep{tpc} tackle this problem by incorporating a dynamic model and applying contrastive learning in the temporal dimension, which encourages the model to capture controllable and predictable information in the latent space. Bisimulation metrics method such as DBC \citep{dbc} and PSE \citep{pse} is another type of representation learning robust to visual distractions. Using the bisimulation metrics that quantify the behavioral similarity between states, these methods make the mode robust to task-irrelevant information. However, DBC cannot generalize to unseen backgrounds \citep{tpc}, and PSE is only shown to work in the model-free setting and requires a pre-trained policy to compute the similarity metrics, while our model learns both the world model and the policy from scratch.

\section{Conclusion}
In this work, we presented the first reconstruction-free MBRL agent based on the prototypical representation and its temporal dynamics. In experiments, we demonstrated the consistently improved accuracy and robustness of the proposed model in comparison to the Temporal Predictive Coding (TPC) agent and the Dreamer agent for both standard and natural background DMC tasks. Our results suggest that there are unexplored broad areas in reconstruction-free MBRL. Interesting future directions are to apply this model on Atari games and to investigate the possibility of learning hierarchical structures such as skills without reconstruction.

\subsubsection*{Acknowledgments}
This work was supported by Electronics and Telecommunications Research Institute (ETRI) grant funded by the Korean government [21ZR1100, A Study of Hyper-Connected Thinking Internet Technology by autonomous connecting, controlling and evolving ways].
The authors would like to thank Tung Nguyen for help with implementing \tpc, Jindong Jiang and Ishani Ghose for editing the related work section, and Chang Chen, Parikshit Bansal, and Jaesik Yoon for fruitful discussion.


\bibliography{iclr2022_conference,refs_ahn}
\bibliographystyle{iclr2022_conference}


\appendix
\section{Hyperparameters}

For hyperparameters that are shared with $\dreamer$, we use the default values suggested in the config file in the official implementation of $\dreamer$, with the following two exceptions. We set \texttt{rssm.discrete = False} and \texttt{actor.dist = tanh\_normal}, as we find these changes improve performance over the default setting. The additional hyperparameters introduced in $\dreamerp$ are listed in Table~\ref{tbl:param}. We find it helpful to freeze the prototypes for the first $10$K gradient updates. In the natural background setting, we add a squared loss that encourages the $\ell_2$-norm of projections (before $\ell_2$-normalization) to be close to $1$. This helps stabilize the model.

\begin{table}[ht]
	\centering
	\caption{Additional hyperparameters in $\dreamerp$.}
	\label{tbl:param}
	\vskip 0.1in
	\begin{adjustbox}{max width=\textwidth}
  	\begin{tabular}{cc}
      \toprule
      Hyperparameter & Value \\
      \midrule
  	  \multicolumn{1}{l}{Number of prototypes $K$} & $2500$ \\
  	  \multicolumn{1}{l}{Prototype dimension} & $32$ \\
  	  \multicolumn{1}{l}{Softmax temperature $\tau$} & $0.1$ \\
      \multicolumn{1}{l}{Sinkhorn iterations} & $3$ \\
      \multicolumn{1}{l}{Sinkhorn epsilon} & $0.0125$ \\
      \multicolumn{1}{l}{Momentum update fraction $\eta$} & $0.05$ \\
      \bottomrule
  	\end{tabular}
  	\end{adjustbox}
\end{table}

\end{document}